# OntoChatGPT Information System: Ontology-Driven Structured Prompts for ChatGPT Meta-Learning

**OLEKSANDR PALAGIN[1], VLADISLAV KAVERINSKY[2], ANNA LITVIN[1], KYRYLO MALAKHOV[1]**

[1]Microprocessor technology lab, Glushkov Institute of Cybernetics of the National Academy of Sciences of Ukraine, Kyiv, Ukraine

[2]Department of Wear-Resistant and Corrosion-Resistant Powder Construction Materials, Frantsevic Institute for Problems in Material Science of the National Academy of Sciences of Ukraine, Kyiv, Ukraine

Corresponding author: Oleksandr Palagin (e-mail: palagin_a@ukr.net).

This work was supported by the National Research Foundation of Ukraine under grant agreement – "Development of the cloud-based platform for patient-centered telerehabilitation of oncology patients with mathematical-related modeling", application ID: 2021.01/0136.

**ABSTRACT** This research presents a comprehensive methodology for utilizing an ontology-driven structured prompts system in interplay with ChatGPT, a widely used large language model (LLM). The study develops formal models, both information and functional, and establishes the methodological foundations for integrating ontology-driven prompts with ChatGPT's meta-learning capabilities. The resulting productive triad comprises the methodological foundations, advanced information technology, and the OntoChatGPT system, which collectively enhance the effectiveness and performance of chatbot systems. The implementation of this technology is demonstrated using the Ukrainian language within the domain of rehabilitation. By applying the proposed methodology, the OntoChatGPT system effectively extracts entities from contexts, classifies them, and generates relevant responses. The study highlights the versatility of the methodology, emphasizing its applicability not only to ChatGPT but also to other chatbot systems based on LLMs, such as Google's Bard utilizing the PaLM 2 LLM. The underlying principles of meta-learning, structured prompts, and ontology-driven information retrieval form the core of the proposed methodology, enabling their adaptation and utilization in various LLM-based systems. This versatile approach opens up new possibilities for NLP and dialogue systems, empowering developers to enhance the performance and functionality of chatbot systems across different domains and languages.

## I. INTRODUCTION

Automatic dialogue systems have been developed for several decades, ever since the advent of computers with user interfaces. The concept of utilizing natural language for human-machine interaction has always been highly desirable. It offers convenience and ease compared to the need to learn a specific language and follow predefined instructions. The introduction of textual interfaces, followed by windows and menu-based interfaces, marked a significant breakthrough, making computers accessible and widely used tools for various users.

However, despite their practicality and convenience, these interfaces still lack the necessary flexibility. They are often characterized by rigid predetermined structures and can become complex and intricate. Consequently, users are required to invest significant time and effort in familiarizing themselves with all the features of such interfaces. It would be ideal if users could simply express their desired actions or requested information in a natural language, either through speech or typing, thereby eliminating the need for extensive interface exploration.

By incorporating natural language understanding and processing capabilities into dialogue systems, users would benefit from a more intuitive and user-friendly interaction. This advancement would enhance the efficiency and usability of these systems, ensuring a smoother user experience and reducing the learning curve typically associated with complex interfaces.

Currently, there exists a range of virtual assistants that employ natural language processing both in written and spoken form. Prominent examples include AI systems like Apple Siri, Google Assistant, Amazon Alexa and Microsoft Cortana, among others. However, the development of ChatGPT (which is based upon OpenAI's GPT-3, GPT-3.5 and GPT-4 foundational GPT models [1], and has been fine-tuned for conversational applications using both supervised and reinforcement learning techniques) [2]–[4] has marked a significant breakthrough in the field of artificial intelligence, particularly in the domains of natural language processing (NLP) and understanding (NLU). The "GPT" stands for generative pre-trained transformer – a type of large language model (LLM). ChatGPT is a purely textual system and lacks the ability to recognize and generate oral speech or interact with physical objects in the material world. Nevertheless, its potential as a powerful natural language system is vast, offering a wide range of capabilities for information provision, generation, and structuring. Various program features and tricks can be accomplished using ChatGPT.

Consequently, ChatGPT not only serves as a valuable standalone virtual assistant and companion but also holds great potential for integration into other software systems, leveraging its abilities to fulfill specific purposes. This perspective has opened up new avenues for research, particularly in exploring how ChatGPT can be harnessed to serve determined goals required for system activities.

Researchers now have the opportunity to delve into investigating methods to effectively utilize and optimize ChatGPT's capabilities within specific domains. This includes adapting and customizing ChatGPT to perform tasks and address challenges tailored to the unique requirements of different software systems. By effectively "taming" ChatGPT, researchers can harness its strengths and align it with the specific objectives of various applications, leading to further advancements in the field of natural language processing and expanding the boundaries of what can be achieved through intelligent information systems.

The aim of the research discussed in this paper is to establish formal models (both information and functional), and to develop methodological foundations for utilizing an ontology-driven structured prompts system in interplay with ChatGPT. The system developed in this research is called *OntoChatGPT*. This system enables the provision of information and *inference* (according to the definition given in "The explanatory ontograph dictionary for knowledge engineering"[1] – expanding the knowledge base by deriving new information from existing knowledge units; this process includes various operations, with logical deduction being a notable case) based on a specific set of contexts, functioning as a dialogue system. Logical deduction involves inferring new information based on established facts, rules, and logical principles. It enables the system to draw logical conclusions and make connections between different pieces of information. By employing deductive reasoning, the system can extend its understanding and generate additional knowledge units that were not explicitly provided. This process of obtaining new information units from previously known ones plays a crucial role in enhancing the system's knowledge and improving its overall functionality. It enables the system to make intelligent inferences, uncover hidden relationships, and provide more comprehensive and valuable insights to the user.

The implementation of this technology is demonstrated using the Ukrainian language and applied within the domain of rehabilitation (specifically e-rehabilitation [5]).

By developing formal models, this research provides a structured framework for organizing and representing knowledge in a systematic manner. These models, encompassing both information and functional aspects, lay the foundation for effectively integrating an ontology-driven approach with ChatGPT. The combined OntoChatGPT system enables sophisticated dialogue interactions that incorporate inference and leverage contextual information to provide meaningful responses.

Furthermore, this research focuses on the practical application of the developed technology within the field of rehabilitation. By implementing and testing the OntoChatGPT system using the Ukrainian language, the study demonstrates the potential and versatility of the approach. Specifically, it showcases how the ontology-driven structured prompts system, in conjunction with ChatGPT, can enhance information provision and inference in the context of rehabilitation-related discussions.

This work not only contributes to the advancement of dialogue systems and natural language processing but also demonstrates the applicability and relevance of the proposed methodology within a specific domain. The findings offer valuable insights into the potential benefits of integrating ontology-driven structured prompts systems with state-of-the-art language models, paving the way for further developments and applications in the field of intelligent information systems and knowledge technologies.

## II. RELATED WORK

The ChatGPT does not have a conventional API that consists of a fixed number of URLs and corresponding commands with predetermined actions. However, it does have an API that accepts natural language commands [6]. Nonetheless, there are certain peculiarities associated with this approach. Firstly, while ChatGPT possesses knowledge across a wide range of languages, its understanding and proficiency levels in each language may vary. English serves as the primary language for ChatGPT, and commands and instructions should be written in English, even when dealing with other languages.

Another crucial aspect is that instructions provided to ChatGPT need to be well-structured, clear, and precise. Due to the limitation on the number of tokens that can be processed by ChatGPT, instructions should be concise yet informative. Experimental evidence with ChatGPT [7]–[10] has revealed that one effective approach for delivering concise yet comprehensive commands and instructions is by formatting them as JSON.

By structuring instructions in JSON format, researchers and developers can ensure that the information is organized, easily interpretable, and maximally efficient for ChatGPT. This approach enables clear communication of the desired actions and expectations to the system, optimizing the interaction between users and ChatGPT.

Furthermore, it is essential to strike a balance between brevity and clarity in the instructions provided to ChatGPT. While instructions should be concise to accommodate token limitations, they must also convey sufficient information to instruct ChatGPT accurately. Achieving this balance ensures that the system's responses align with the user's intentions and expectations.

Despite the absence of a traditional API, ChatGPT offers an API that accepts natural language commands. Adhering to English as the main language, structuring instructions in a clear and precise manner, and leveraging JSON formatting for concise yet comprehensive commands are key considerations for effectively utilizing ChatGPT's capabilities. These strategies enhance the interaction between users and the system, facilitating more accurate and meaningful responses, compared with chain-of-thought reasoning technique [9], [11]–[14].

---

[1] https://www.dropbox.com/s/kg4w2rfluij3tuy/expl-onto-dict-ke.pdf?dl=0

In public GitHub repository "Mr. Ranedeer: Your personalized AI Tutor!" [7], an example of an instruction set aimed at transforming ChatGPT into a virtual tutor can be found. These instructions are formatted as nested dictionaries, with concise key terms representing the main concepts of the intended purpose. The values associated with these keys can be dictionaries, providing further details, or natural language (English) phrases offering comprehensive and clear explanations.

To utilize this virtual tutor functionality, one can simply copy and paste the provided instruction into the ChatGPT interface. By doing so, ChatGPT can be adapted to serve as a virtual tutor across various subject areas covered in its knowledge base. The concept of using structured prompts to instruct ChatGPT's behavior in a desired manner is both tempting and promising.

Furthermore, ChatGPT has the capability to incorporate plugins stored in external resources [15]–[17]. Links to these resources can also be included in the instructions, thereby expanding the range of functionalities and possibilities. This opens up new avenues of research, referred to as prompt engineering and meta-learning. Recent studies [11], [13], [18]–[20] highlight the significance and relevance of exploring these areas.

The main objective of prompt engineering is to address the challenge of guiding ChatGPT towards appropriate responses, particularly in tasks requiring logical derivations. Instructions can be crafted to clarify the task's intricacies and break it down into sequential steps, guiding the AI towards the desired outcome. Prompt engineering is akin to an art form, involving the careful selection of specific words, phrase structures, and their order to elicit the desired AI behavior. Various strategies have been developed, including the use of imperatives to define the AI's role, planned sequences, structured data formats (such as JSON, XML, YAML), self-critique chains, and others.

Determining the most effective strategy for prompt engineering remains an open question, but promising approaches have been reported in [10], [21]. This work emphasizes the importance of prompt phrase structure and the utilization of specific words and expressions. Combining the findings from this study with other relevant research can yield valuable insights and contribute to advancing the field.

Prompt engineering opens up possibilities for providing targeted and specific learning to ChatGPT, enabling it to gain a deeper understanding of subjects it may not have sufficient knowledge about. While mechanisms like model training and fine-tuning exist in ChatGPT, they can be costly and require large, carefully curated datasets. In some cases, it may not be feasible or practical to follow this approach. Instead, valuable information can be conveyed in textual form or through data structures combined with JSON prompts that instruct ChatGPT algorithms on how to process the provided data. This allows for the expansion of ChatGPT's knowledge and capabilities, making it suitable for dialogue systems or even control systems.

While employing a rigid structure like the one described in [22] may be a functional approach, there is room for further development and exploration. Systems that interact with ChatGPT can utilize a variety of instructions or templates tailored for different purposes. The prompts themselves can be made more flexible by incorporating optional fields and providing different explanations (values) for each field. Such a system should include instructions on when and how to use the templates with ChatGPT and what specific values should be used in different cases. These instructions for creating and utilizing structured prompts in ChatGPT can be organized within an ontology, resulting in an ontology-driven system.

The utilization of ontologies, or meta-ontologies, as a repository of system behavior rules is discussed in [21], albeit without direct reference to chat ChatGPT or similar applications. In this approach, the ontology serves as a decision-making module, guiding the system on how to handle specific data types and represent them in the user interface.

By incorporating ontologies to instruct the behavior of ChatGPT and leveraging structured prompts, we can develop a powerful ontology-driven system. This system enhances ChatGPT's ability to adapt and learn in specific domains, leveraging the flexibility of prompts, and benefiting from the knowledge stored within the ontology. The combination of prompt engineering, ontology utilization, and ChatGPT-based chat systems holds great potential for advancing intelligent information systems and knowledge technologies in various domains.

The foundational concepts of information systems with an *ontology-driven architecture* are extensively discussed in [23], [24]. In "The explanatory ontograph dictionary for knowledge engineering", *ontology-driven architecture* – is defined as a system architecture that revolves around two main components: an "Active" computer ontology and a "Problem Solver". These components work collaboratively to govern the information processing process, with a specific focus on addressing practical user problems and supporting targeted activities. Furthermore, an *ontology-driven information system* is characterized as a comprehensive system comprising several key elements. These include a knowledge base that is intricately linked to ontologies (typically represented as a finite collection of systematically integrated knowledge bases within specific subject domains), an inference engine, an application processing subsystem, and interfaces (UI, API) for user interaction and/or external environment integration. Collectively, these components facilitate the effective usage of ontological knowledge within the information system.

The approach we are adopting in this work is distinct from the methodology employed in our previous researches [25], [26]. In our previous works, the ontology served as the primary repository for information within the dialogue system, rather than as a container for rules, which were organized using a different approach. Nonetheless, certain elements from those previous developments are still applicable to our current endeavor. For example, techniques such as named entity extraction, linked context analysis, and the automatic generation of formal SPARQL [27] queries from user-provided natural language phrases are utilized. Addressing these challenges is crucial and inevitable in the development of such systems [28]–[30]. However, they fall outside the scope of the present work.

In our previous studies [25], [26], we focused on utilizing the ontology as a central storage for the information exchanged within the dialogue system. In contrast, the current work explores the utilization of the ontology as a framework for defining rules and guiding the behavior of the system. Despite this shift in approach, certain aspects of our previous developments remain relevant. Specifically, techniques such as named entity extraction, linked context analysis, and the automatic generation of formal SPARQL queries from user-provided natural language phrases have proven valuable and are also incorporated into the current work.

However, it is important to note that the challenges associated with these techniques, such as ensuring accurate entity extraction and generating precise SPARQL queries, are complex and require dedicated research efforts [28]–[30]. While these topics are critical for system development, they lie beyond the scope of the present study.

In summary, while our previous works [25], [26] employed ontology as a primary information repository within the dialogue system, the present work utilizes the ontology as a means of defining rules. Despite this shift, certain techniques from our earlier research, including named entity extraction, linked context analysis, and SPARQL query generation, continue to be relevant. However, addressing the challenges associated with these techniques remains an ongoing focus of research, which extends beyond the scope of the current study.

## III. FORMAL MODELS

### A. INFORMATION MODEL OF THREE-TUPLE COMPOSITE SERVICE – OntoChatGPT INFORMATION SYSTEM

The OntoChatGPT information system utilizes an information model based on a three-tuple composite service. This information model forms the foundation of OntoChatGPT's functionality and allows for the integration of diverse services within the system.

The OntoChatGPT information system is represented as a three-tuple composite service (CS) using the revised formalisms given in [31], [32]:

$$CS_{OntoChatGPT} = \langle D_{evkit}, F_{unc}, E_{nv} \rangle \qquad (1)$$

where:

$D_{evkit} = \{ws_w, as_d \mid w = \overline{1,k}, d = \overline{1,l}\}_{k,l \in \mathbb{N}}$ – is a comprehensive set of web services and application software available for developers which and enables the development of various applications and services within the system – OntoChatGPT development kit. $\mathbb{N}$ denotes a set of nonnegative integer numbers.

The formalization of the web service, denoted as $ws$, is an extension of the $Service$ formalism introduced in [33]. This specialized representation incorporates additional properties, namely $m_{read}$, $h_{read}$ and $r_{est}$, which enhance the descriptive power and characteristics of the formal model:

$$ws = \{p_{re}, e_{ff}, i_{nput}, o_{utput}, p_{rovider}, c_{aller}, d_{esc}, r_{est}\} \qquad (2)$$

where:

$c_{aller}$ – *caller* is the consumer or user of the web service.

$p_{re}$ – in the context of web services $ws$, *preconditions* refer to the conditions that must be satisfied before a web service can be consumed. They define the prerequisites that need to be met by the caller $c_{aller}$ before invoking the service.

$e_{ff}$ – *effects* represent the conditions or changes in the world that can be expected to be true after the web service $ws$ has been executed. They indicate the outcomes or results of performing the service. Within the preconditions $p_{re}$ and effects $e_{ff}$ framework, there are special subclasses known as input $i_{nput}$ and output $o_{utput}$.

$i_{nput}$ – *input* conditions correspond to preconditions $p_{re}$, specifying the necessary input data or parameters required by the web service $ws$.

$o_{utput}$ – *output* conditions, on the other hand, align with effects $e_{ff}$, denoting the expected output or outcomes produced by the web service $ws$.

$p_{rovider}$ – is the *provider* entity responsible for offering the web service $ws$.

$d_{esc} = \{m_{read}, h_{read}\}$ – is a *description* of the particular web service $ws$ is provided in both machine-readable $m_{read}$ and human-readable $h_{read}$ formats. This description, known as $d_{esc}$, serves as a resource accessible to the caller $c_{aller}$, providing information about the web service $ws$ and its functionality.

Additionally, the creation of web services $ws$ adheres to a set of constraints $r_{est}$, influenced by the RESTful architectural style as outlined in [34]. These constraints include:

- *Client/Server*: This constraint emphasizes the separation of concerns by adopting a client-server architecture. It allows for independent evolution of different components, enabling the client's user interface to evolve separately from the server and promoting simplicity in the server's design.
- *Stateless*: The client-server interaction is designed to be stateless, meaning that the server does not store any client-specific context. Instead, the client maintains any necessary session information, ensuring that each request can be treated independently.
- *Cacheable*: Data within a response can be labeled as cacheable or non-cacheable. If a response is cacheable, the client or intermediary can reuse it for similar future requests, reducing the need for redundant interactions with the server.
- *Uniform Interface*: The uniform interface constraint ensures that there is a consistent and standardized interface between components. This uniformity facilitates interoperability and allows clients, servers, and network-based intermediaries to depend on the predictability of the interface's behavior.
- *Layered System*: Components are organized into hierarchical layers, where each component is only aware of the layer with which it directly interacts. This layered approach promotes modularity and scalability, as components can operate within their designated layers without requiring knowledge of other layers.

- *Code on Demand*: This constraint is optional and provides support for extending client functionality through the downloading and execution of scripts. Clients can dynamically enhance their capabilities by acquiring and running code components from the server.

By adhering to these constraints, the RESTful architectural style offers a framework for creating web services that are modular, scalable, stateless, cacheable, and exhibit a uniform interface.

The formalization of application software $as$, encompassing both desktop applications and utilities that feature graphical or command-line user interfaces, can be considered a specialization within the broader *Service* formalism discussed in [33]. In this specialized context, an additional property $d_{esc}$ is introduced, denoted by the human-readable $h_{read}$ description of the particular desktop application:

$$as = \{p_{re}, e_{ff}, i_{nput}, o_{utput}, p_{rovider}, c_{aller}, d_{esc}\{h_{read}\}\} \quad (3)$$

where:

$d_{esc} = \{h_{read}\}$ – is a human-readable $h_{read}$ description of a particular desktop application service, accessible for the caller $c_{aller}$. All other elements in the formalization remain the same as described in equation (2). The elements such as preconditions $p_{re}$, effects $e_{ff}$, input $i_{nput}$, output $o_{utput}$, provider $p_{rovider}$, and caller $c_{aller}$ continue to hold their respective meanings and definitions as previously stated.

Additionally, in the context of application software $as$ formalization, there is no specific set of constraints imposed. Unlike web services $ws$, which adhere to the RESTful architectural style with a defined set of constraints [33], application software $as$ does not have a predetermined set of constraints that govern its design and behavior. Instead, the constraints applicable to application software $as$ may vary depending on the specific requirements, platform, and design principles employed during its development. Therefore, the formalization of application software $as$ allows for greater flexibility and adaptability, as it can encompass a wide range of applications with different constraints and design considerations. This flexibility enables developers to tailor the software to meet the unique needs of users and provide a seamless user experience, whether through a graphical or command-line interface.

$F_{unc} = D_{evkit} : \left\{ C_j \mid j = \overline{1,n} \right\}_{n \in \mathbb{N}}$ – is a set of functions that encompass the functional aspects of OntoChatGPT's information technology. Each function corresponds to a specific knowledge management pipeline or process, which arises from the integration and interaction of the elements within the $D_{evkit}$. $\mathbb{N}$ denotes a set of nonnegative integer numbers.

$C_j \subseteq D_{evkit}, C_j = \left\{ ws_o, as_p \mid o, p \geq 0, o \leq k, p \leq l \right\}_{o, p \in \mathbb{N}}$

– is a subset of web services and application software that are required for the successful implementation of the *j*-th function within the $D_{evkit}$. This subset specifically caters to the requirements of the respective function. $\mathbb{N}$ denotes a set of nonnegative integer numbers.

$E_{nv} = \{prl, os, floss\}$ – is a set of elements that come together as layers forming the *Knowledge Integrated Development Environment* (K-IDE). Each element within this set contributes to the overall functionality and capabilities of the K-IDE.

The element $prl$, which stands for the physical resource layer, represents the physical hardware and facility resources as defined in [35]. It encompasses the tangible components that form the foundation for the K-IDE infrastructure. The $prl$ layer ensures the availability and proper functioning of the necessary physical resources required to support the K-IDE framework.

The element $os$, which refers to the operating system layer, represents the guest operating system within the K-IDE. The operating system layer is designed to utilize Unix-like operating systems, such as Ubuntu Server for x86 systems and DietPi Debian-based lightweight operating system for ARM-based single board computers. It supports various light-weight desktop environments including LXDE, XFCE, or LXQt. This layer provides the foundation for running the K-IDE framework and ensures compatibility with the selected operating system environments and desktop environments.

The FLOSS layer, denoted as $floss = \{in, ex\}$, represents the Free/Libre and open-source software (FLOSS) component within the K-IDE. This layer encompasses both internal software components, represented by the set $in = \left\{ ws_w, as_d \mid w = \overline{1,k}, d = \overline{1,l} \right\}_{k, l \in \mathbb{N}}$, and external software components, represented by the set $ex = \{ws_i\}, i \in \left\{\overline{1,n}\right\}, n \in \mathbb{N}$. $\mathbb{N}$ denotes a set of nonnegative integer numbers.

The internal software components $in$ include a comprehensive application suite tailored for the scientific research and development lifecycle, along with additional application software $as$ and web services $ws$. On the other hand, the external software components $ex$ refer to specific web services $ws$.

It is important to note that the $D_{evkit}$ subset is part of the FLOSS layer $D_{evkit} \subset floss$, signifying that the development kit is built upon and aligned with the principles of Free/Libre and open-source software.

### B. OntoChatGPT DEVELOPMENT KIT

In the current stage of OntoChatGPT information technology, the development kit $D_{evkit}$ set consists of the following comprehensive collection of problem-oriented web services $ws$ and application software $as$:

$ws_1$ – WebProtégé [36] – is an external web service $ws_1 \in ex$, $ws_1 \in D_{evkit}$. It serves as a free and open-source ontology development environment designed for the Web. With WebProtégé, users can effortlessly create, upload, modify, and collaborate on ontologies, enabling seamless collaborative viewing and editing experiences.

$ws_2$ – Apache Jena Fuseki [37] – is a FLOSS that provides an HTTP interface for working with RDF data. Fuseki is a part of the Apache Jena Java framework and offers robust support for SPARQL, enabling seamless querying and updating of RDF data through its SPARQL server engine [38]. Fuseki can be locally deployed within the K-IDE environment $E_{nv}$ as an internal component $ws_2 \in in$, or it can be externally deployed via the Software-as-a-Service application delivery model (SaaS) $ws_2 \in ex$ [35], also $ws_2 \in D_{evkit}$.

$ws_3$ – KEn (former Konspekt)[2] – is an NLP-powered network toolkit (web service with API) for contextual and semantic analysis with document taxonomy building feature. The KEn web service supports processing of English, Ukrainian and Norwegian (Bokmal). The KEn web service offers comprehensive coverage of essential stages in NLP. These stages include: text data extraction; text preprocessing, spell checking and automatic correction, sentence/word tokenization, part-of-speech tagging, lemmatization, word stemming, shallow parsing, JSON/XML-structure generation.

KEn web service can be deployed locally as a part of K-IDE $E_{nv}$, as $ws_3 \in in$, or can be deployed externally as $ws_3 \in ex$ via SaaS [35], $ws_3 \in D_{evkit}$.

$ws_4$ – natural language phrase analysis network service [25] – is a specialized web service that supports natural language text in both Ukrainian and English, enabling the construction of semantic trees for phrases. These semantic trees is a key part in facilitating SPARQL queries to form connections with ontologies. Each semantic tree is defined by marker words and expression types, providing valuable insights into the structure of the sentence. In certain cases, multiple semantic trees can be identified within an initial sentence, allowing for the generation of suitable SPARQL queries for each specific tree. This web service can be deployed locally as a part of K-IDE $E_{nv}$, as $ws_4 \in in$, or can be deployed externally as $ws_4 \in ex$ via SaaS [35], $ws_4 \in D_{evkit}$.

$ws_5$ – OpenAI ChatGPT Playground [39] – is an interactive web-based platform that allows users to experiment with the capabilities of the ChatGPT language model. It provides a user-friendly interface where individuals can input text prompts and receive responses generated by ChatGPT in real-time. The Playground offers a range of features to enhance the user experience, including options to adjust the model's temperature and sampling settings. Playground is an external web service $ws_5 \in ex$, $ws_5 \in D_{evkit}$.

$ws_6$ – UkrVectōrēs (former docsim)[3] – an NLU-powered tool for knowledge discovery, classification, diagnostics and prediction. UkrVectōrēs can be described as a "cognitive-semantic calculator" that serves as a powerful tool for distributional analysis. This web service encompasses several essential elements, including: semantic similarity calculation (UkrVectōrēs allows for the computation of semantic similarity between pairs of entities; this feature provides insights into the relatedness and proximity of words in a semantic space); word nearest neighbors (this functionality aids in exploring words with similar meanings or associations); algebraic operations on word vectors (UkrVectōrēs supports various algebraic operations on word vectors, such as addition and subtraction); semantic mapping (users can generate semantic maps that depict the relations between input words; these maps are valuable for visualizing clusters, oppositions, and exploring hypotheses related to semantic relationships); access to raw vectors and visualizations features; use of third-party prognostic models.

$as_1$ – Apache Jena ARQ [37] is a SPARQL query engine Java-based command-line utility. ARQ is a part of FLOSS Java framework Apache Jena. The main ARQ features are: SPARQL 1.1 support; client-support for remote access to any SPARQL endpoint (including usage of SPARQL 1.1 SERVICE keyword); support for federated query; access and extension of the SPARQL algebra. $as_1 \in in$, $as_1 \in D_{evkit}$.

$as_2$ – nlp_api [40] – is a collection of scripts (NLP API from Language Tool) designed for essential text preprocessing tasks specifically tailored to Ukrainian language. $as_2 \in in$, $as_2 \in D_{evkit}$.

$as_3$ – is a desktop application service that enables the semi-automatic and fully automatic generation of an OWL ontology [25] from natural language text. It also supports the semi-automatic import of knowledge from a dataset, capturing it as RDF triples, and storing it in an RDF triplestore (TDB or TDB2 component of Apache Jena for RDF storage and query [38]; Apache Jena Fuseki) or in the graph database (Neo4j) [38].

### C. FUNCTIONAL MODEL OF THE OntoChatGPT INFORMATION SYSTEM

The functional enrichment of the OntoChatGPT information system is represented by the following set $F$ of functions synthesized from the $D_{evkit}$:

$$F = \{C_1, C_2, C_3, C_4\} \qquad (4)$$

where:

$C_1$ – semi-automatic import of knowledge from a dataset and capturing it as RDF triples snapshot in RDF triplestore.

$C_2$ – semi-automatic and fully automatic generation of an OWL ontology from natural language text.

$C_1$ and $C_2$ functions: expanding beyond the scope of this research. Please note that a comprehensive description of the $C_1$ and $C_2$ functions, along with their respective information and functional models, can be found in our previous articles [25], [26]. For the purpose of this article, we will focus on $C_3$ and $C_4$ functions of the OntoChatGPT information system.

---
[2] https://github.com/malakhovks/ken

[3] https://github.com/malakhovks/docsim

$C_3$ – ontology-driven dialogue function that integrates ChatGPT and the structured prompts.

$C_4$ – structured prompts for ChatGPT meta-learning function.

In the next section, we delve into a comprehensive study of the functional models and methodological foundations underlying two key components: $C_3$ and $C_4$.

## IV. METHODOLOGICAL FOUNDATIONS FOR LEVERAGING THE OntoChatGPT INFORMATION SYSTEM

The presented methodology can be divided into two key components, each serving a distinct purpose in the development of the OntoChatGPT system. Firstly, we focus on the technique of prompts-based meta-learning and the creation of structured prompts for ChatGPT. This approach involves leveraging prompts to instruct the meta-learning process of ChatGPT, enabling it to generate more contextually relevant and accurate responses. We delve into the methodology behind designing and implementing these prompts, highlighting their significance in enhancing the conversational capabilities of ChatGPT.

idea behind this system is to incorporate specific subject areas and their associated contexts, which may contain domain-specific information not fully covered in ChatGPT's knowledge base. These contexts are stored in a database, such as MongoDB or a relational database model, and are linked to sets of named entities with their own ontology-like structure. Additionally, sentiment analysis can be used to categorize the contexts. The binding of named entities to their corresponding contexts includes semantic components that elucidate the entity's role within the context. These additional features aim to improve the relevance and clarity of the selected context for subsequent processing. To automate these processes, we utilize our previously developed tools [25] and incorporate transformer pre-trained BERT-based models like [41].

For semantic analysis and named entity extraction from user-provided phrases, ChatGPT proves to be a valuable resource. Special prompts are created specifically for this purpose. Furthermore, ChatGPT is utilized for intent analysis of user phrases. The defined intents, along with extracted named entities annotated with their semantic roles, and the selected list of contexts are then provided as input to ChatGPT. Accompanying these inputs are the appropriate structured

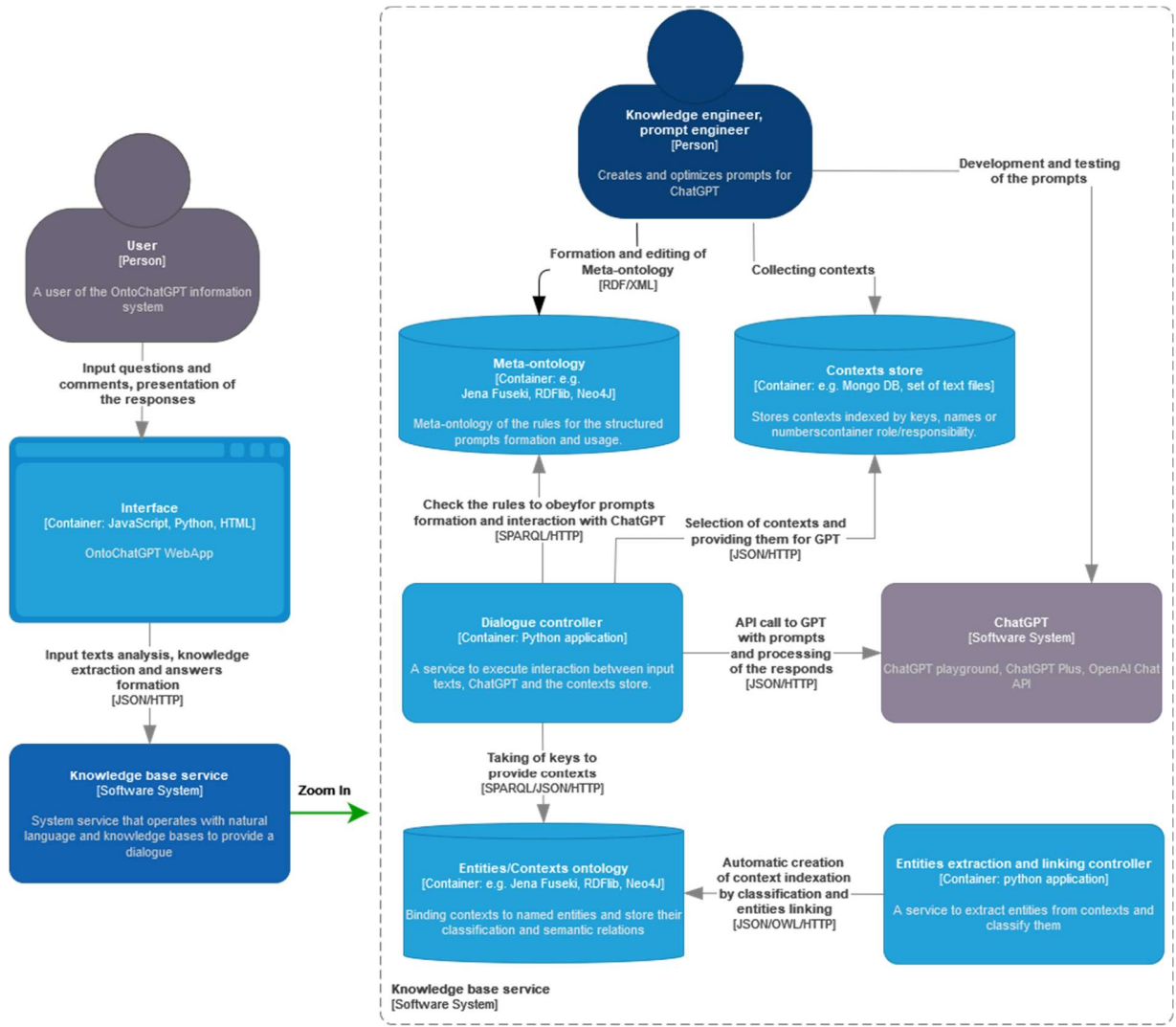

Figure 1: Context/container C4 model diagram of OntoChatGPT information system.

The second part of the methodology centers around the development of an automatic ontology-driven dialogue system that integrates ChatGPT and the structured prompts. The core prompts that clarify the information to be extracted and the desired representation format. To provide a visual representation of the overall system scheme, we present a context/container C4 model diagram [42] as depicted in Figure 1. This diagram offers a comprehensive overview of the system architecture, showcasing the interplay between various components and their relationships. It serves as a visual aid in understanding the underlying structure and functionality of the OntoChatGPT system.

One of the most outstanding features of the system is the flexibility of its structured prompts for ChatGPT. Instead of rigid prompts, they are dynamically generated based on the specific situation using instructions provided in the form of a meta-ontology. This meta-ontology outlines the fields to be included in the JSON (or XML) structure and the corresponding prompt phrases to be inserted. Each instruction or structured prompt for ChatGPT has its own set of fields and predefined values that can be incorporated. Additionally, the prompt includes a template structure for the response, which ensures consistency and simplifies subsequent processing. The creation of prompt phrases uses proven techniques from [11], [18] to achieve effective and coherent prompts.

The operation of the OntoChatGPT system can be conceptualized through a Petri Net-like scheme, specifically in the form of a modified System Net Marking Graph [43]. This formalization provides a structured representation of the system's functioning, capturing the flow and interactions between various components. A more detailed formulation is given in Eq. (5), (6).

$$C_3 = M_0 \xrightarrow{t,text} M_1 \begin{Bmatrix} \xrightarrow[\text{INT-DEF}]{p_1,O_{meta},text'} M_2 \xrightarrow[\text{INF-PR-FORM}]{O_{meta},s_{in}} M_3 \\ \xrightarrow[\text{CON-INT-DEF}]{p_2,O_{meta},text'} M_4 \xrightarrow[\text{CON-PR-FORM}]{O_{meta},d} M_5 \\ \xrightarrow[\text{ENT-EXTR}]{p_3,O_{meta},text'} M_6 \end{Bmatrix} \xrightarrow{O_{con},e,s_{in}} M_7 \begin{Bmatrix} \xrightarrow[\text{INF-EXTR}]{p_{4_1},s_{in_1},c} M_8 \\ \xrightarrow[\text{INF-EXTR}]{p_{4_2},s_{in_2},c} M_8 \\ \dots \\ \xrightarrow[\text{INF-EXTR}]{p_{4_n},s_{in_n},c} M_8 \\ \xrightarrow[\text{CON-DER}]{p_5,d,c} M_9 \end{Bmatrix} \xrightarrow[\text{RES-FORM}]{r,O_{meta}} M_{10} \quad (5)$$

In the OntoChatGPT system framework, several components, states, processes, and variables contribute to the functionality and operation of the dialogue system. Here is an overview of the key elements.

$C_3$ component (is shown in equation (5)) – ontology-driven dialogue function that integrates ChatGPT and the structured prompts. Represents the process of the dialogue act between the user and the OntoChatGPT system. Where:

*States* ($M_0$ to $M_{10}$):

$M_0$ – initial state of receiving a text message from the user.

$M_1$ – pre-processed text state ready for subsequent operations.

$M_2$ – state of the defined intents list.

$M_3$ – state of the formed information extraction prompt template.

$M_4$ – state of an inference to be made.

$M_5$ – state of the formed inferences derivation prompt template.

$M_6$ – State of the extracted named entities list.

$M_7$ – state of the selected contexts list.

$M_8$ – state of the extracted information from the contexts.

$M_9$ – state of inferences derived from the contexts.

$M_{10}$ – state of the results obtained and presented in a suitable form for the user.

*Processes:*

PREP – initial text pre-processing.

INT-DEF – defining the intents expressed in the input text.

CON-INT-DEF – identifying if the intent involves deriving inferences, prompting ChatGPT to provide related information from the passed contexts or related knowledge.

ENT-EXTR – extraction of named entities from the input text.

INF-PR-FORM – formation of prompts for defining intents.

CON-PR-FORM – formation of a prompt to generate inferences.

CX-SEL – selection of relevant contexts based on identified named entities and intents.

INF-EXTP – information extraction from the selected contexts based on the intent.

CON-DER – derivation of inferences from the selected contexts.

RES-FORM – formatting and representation of the results to the user.

*Variables:*

$t$ – set of rules and operations to be applied during the initial input text preprocessing.

$text$ – raw input text provided by the user.

$text'$ – pre-processed and cleaned text for subsequent operations.

$O_{meta}$ – meta-ontology that includes operating rules and prompt formation instructions.

$O_{con}$ – contexts ontology related to the system.

$p_1$ – prompt for defining intents.

$p_2$ – prompt to analyze if any conclusions are to be made.

$p_3$ – prompt for named entity extraction.

$p_{4_n, n \in \mathbb{N}}$ – prompts for information extraction based on a specific intent $n, n \in \mathbb{N}$.

$p_5$ – prompt for ChatGPT to generate conclusions from the given contexts.

$s_{in}$ – set of intents $in$ defined in the input text, activated with specific entities.

$s_{in_n, n \in \mathbb{N}}$ – a specific single intent $in_n, n \in \mathbb{N}$.

$d$ – indicates whether and which conclusions are expected to be made.

*e* – set of named entities found in the input text.

*c* – selected contexts.

*r* – data structures representing the results obtained from ChatGPT.

The dialogue act procedure can be outlined as follows: Upon receiving textual information (request) from the user, it undergoes intent analysis using ChatGPT with the assistance of a specific prompt. The outcome is a list of dictionaries containing the following keys:

- "name": the name of an intent from the provided list in the prompt;
- "type': a more general classification of the intent, such as narration, interrogation, or imperative;
- "probability": a float value ranging from 0 to 1, indicating the probability of the intent being present in the user's text;
- "subject": the subject associated with the intent, if applicable;
- "object": the object associated with the intent, if applicable.

$$C_4 = M_0 \xrightarrow[GUESS]{purpose} M_1 \xrightarrow[ChatGPT]{p} M_2 \left\{ \begin{array}{c} \xrightarrow[CORR]{res,p'} M_3 \\ M_5 \xleftarrow[ChatGPT]{res'} M_3 \end{array} \right\} \xrightarrow[ONT]{p_{fin}} M_5 \qquad (6)$$

The prompt can include various possible intents, such as "quantity", "way of doing", "object", "subject", "action", "location", "direction", "scene of action", "conditions", "instrument", "collaborator", "relation", "cause", "sequence", "origin", and others. These intents represent semantic categories and provide a framework for understanding the user's request.

Additionally, the structured prompt includes fields for information to provide, language to use, and other technical details related to input and output.

Simultaneously, the extraction of named entities is performed using ChatGPT with the aid of another prompt. The result should provide lemmatized words grouped according to entities, specify the type of each group (name or verbal), and indicate the main word within each group.

Based on the extracted named entities and their semantic roles (from intents), the system selects the appropriate contexts from the contexts ontology.

Next, a request is made to ChatGPT, incorporating the previously selected contexts and intent lists, along with the relevant entities (subjects and objects) actualized within the prompts. The form of this prompt can vary significantly and depends on the specific intent or list of intents. The rules for its composition, including the required fields and prompt phrases, are defined in the meta-ontology.

The meta-ontology also encompasses rules for the final representation of results, depending on the types of fields involved. This may include plain text, numbers, dates, lists, tables, or links to external resources, among other possibilities.

The creation of the meta-ontology involves meta-learning, which encompasses the development and fine-tuning of appropriate prompts. This iterative process involves a knowledge engineer (an individual actor) and ChatGPT's Playground. Initially, the knowledge engineer formulates a structured prompt in JSON (XML, YAML, etc.) format, consisting of keys and prompt phrases designed with a specific purpose, drawing from past experience in prompt development and general knowledge. This prompt is then provided to ChatGPT, and the response is analyzed to determine if it adequately satisfies the intended objective.

If the ChatGPT response is completely incorrect or exhibits drawbacks or disadvantages, modifications are made to the initial prompt. These changes may involve adding additional fields, removing redundant or ineffective aspects of the prompt, and editing prompt phrases to enhance task clarity. The improved prompt is then passed back to ChatGPT for further iterations. This iterative process continues until the obtained response closely aligns with the desired and expected result.

Once an appropriate set of prompts has been developed and the behavior of ChatGPT on these prompts has been studied, they are consolidated into the meta-ontology format. The creation of the meta-ontology is a manual process, where the structure of the prompt and the properties of their fields are described based on different anticipated situations and objectives for the problem-solving task. A formalized graphical representation of this meta-learning process $C_3$ component is shown in equation (6).

$C_3$ component – the process of meta-learning. Where:

*States* ($M_0$ to $M_5$):

$M_0$ – the purpose of creating the prompt;

$M_1$ – an initial prompt generated based on guesses and common wisdom;

$M_2$ – ChatGPT's response to the initial prompt;

$M_3$ – a corrected and edited prompt;

$M_4$ – ChatGPT's response to the corrected prompt;

$M_5$ – the meta-ontology with instructions and rules for the new prompt.

*Processes:*

GUESS – making educated guesses and considerations for the new prompt;

GPT – Initialization of ChatGPT;

CORR – correction and editing of the prompt;

ONT – integration of the new prompt into the meta-ontology.

*Variables:*

*purpose* – the underlying idea for creating the new prompt;

*p* – the initial prompt;

$p'$ – the revised prompt;

$p_{fin}$ – the final version of the prompt;

*res* – ChatGPT's response to the initial prompt;

$res'$ – ChatGPT's response to the revised prompt.

## V. RESULTS AND DISCUSSION

A prototype of the proposed system has been developed, incorporating all the main suggested components. Although the structure of the meta-ontology can be complex and intricate, it is not feasible to include it in this document. However, you can find the meta-ontology in the public GitHub repository associated with this paper, along with other related materials. The prompts in the system are formulated based on the rules provided and are represented as JSON structures. The following JSON scheme presents an example prompt, which focuses on intent extraction and their association with relevant entities, if applicable:

```
{
  "information to provide": [
    "define intents",
    "find subjects",
    "find objects"
  ],
  "text": "<A text to be analyzed>",
  "language": "Ukrainian",
  "input information field": "text",
  "possible intents": [
    "quantity",
    "place",
    "way of doing",
    "object",
    "subject",
    "action",
    "location",
    "direction",
    "scene of action",
    "conditions",
    "instrument",
    "collaborator",
    "relation",
    "cause",
    "sequence",
    "origin"
  ],
  "several intents": true,
  "intents probability": true,
  "show intent subject": true,
  "max intents number": 4,
  "intents arrange": "by probability",
  "output format": "JSON",
  "output representation template": {
    "result": [
      {
        "intent": "intent name - string",
        "type": "narration, interrogation or imperative",
        "probability": "float value",
        "subject": "subject of the intent as a name group - string",
        "object": "object of the intent as a name or verb group - string"
      }
    ]
  }
}
```

This is an example of a concise and basic prompt that encompasses multiple fields, allowing for a comprehensive explanation of the main features. One important aspect is the "information to provide" instruction, which declares the primary objectives of the task. The input message to be analyzed is specified in the "text" field. To enhance performance, it is beneficial to specify the language of the input text, such as Ukrainian. However, any other suitable natural language can be used depending on the content being processed by ChatGPT. To instruct ChatGPT and restrict its intent definition capabilities, the desired intents should be explicitly specified in the "possible intents" field. Several Boolean-type fields are provided to offer technical information about the output. In this case, they include "several intents", "intents probability", and "show intent subject". The first field allows for the identification of multiple intents in the given text, the second field enables ChatGPT to estimate the probability of each intent, and the last field instructs ChatGPT to specify the subjects that activate the intents. To prevent an excessive number of intents from being defined, their quantity can be limited (in this paper, up to 4 intents) and sorted by probability. It is always beneficial to define a pattern for the output data structure, specifying the format in which the information should be returned as a result. Therefore, an "output representation template" field has been included.

The final responses provided by ChatGPT, based on the defined intents and the selected contexts, may vary depending on the intents themselves. However, they share a common structure, which is a list of dictionaries with the fields "intent" (the intent name) and "results" (a text or list of texts or other data structures), or simply "none" if there is no relevant information to provide.

The proposed method was tested using the first chapter of the Ukrainian version of the "White Book on Physical and Rehabilitation Medicine in Europe" [44], [45] as the subject area, which served as the basis for constructing the context ontology. The answers obtained from the system were classified into the following categories:

- True Positive $TP$: The response was provided by ChatGPT and it was correct.
- True Negative $TN$: The response was not provided by ChatGPT, indicating that it either acknowledged its lack of knowledge or indicated insufficient information in the texts. This category also includes cases where ChatGPT returned "None" as the response when the appropriate information was indeed absent in the contexts.
- False Positive $FP$: The system attempted to provide a response, but it was incorrect.
- False Negative $FN$: The response was not provided by ChatGPT, even though the correct answer was present in the contexts.

It is important to note that the testing phase excluded questions and phrases from unrelated subject areas that had named entities not present in the context ontology. For such cases, the true negative $TN$ result is guaranteed because no relevant contexts would be selected, and there would be no further processing. Therefore, all the queries included in the testing were formulated to go through all the stages of the proposed approach.

Additionally, it should be emphasized that the proposed approach allows for the possibility of multiple responses to a single question, primarily due to the presence of multiple defined intents. During the evaluation process, all the provided responses were taken into account, regardless of whether they were given in response to the same or different questions.

These considerations ensure a comprehensive assessment of the system's performance and its ability to handle various queries while considering the defined intents and extracting relevant information from the selected contexts.

The testing results are given in Table 1.

**Table 1. The proposed method testing results**

|  | True | False |
|---|---|---|
| **Positive** | 17 | 9 |
| **Negative** | 7 | 1 |

The testing of the proposed method yielded the following values for the standard evaluation metrics:
- Accuracy: 0.7059;
- Precision: 0.6534;
- Recall: 0.9444;
- F1 Score: 0.7724.

These metrics provide a quantitative assessment of the system's performance in terms of its accuracy, precision, recall, and overall effectiveness. The *accuracy* metric represents the proportion of correct answers provided by the system compared to the total number of queries. The *precision* metric measures the system's ability to provide accurate responses among the answers it generates. The *recall* metric indicates the system's capability to retrieve all relevant answers from the available contexts. The *F1 score* combines precision and recall to provide a balanced measure of overall performance.

In addition to the standard metrics, we also considered additional criteria, namely *Precision\** and *Recall\** (Eq. 7, 8). These metrics differ from the standard Precision and Recall in that they treat both true positive $TP$ and true negative $TN$ results as true results, without distinguishing between them. These additional criteria provide a broader evaluation of the system's effectiveness in capturing true answers and identifying relevant information from the contexts. By considering both positive and negative results, we gain a more comprehensive understanding of the system's performance in terms of precision and recall.

$$Precission^* = \frac{TP+TN}{TP+TN+FP} = 0.7273 \qquad (7)$$

$$Recall^* = \frac{TP+TN}{TP+TN+FN} = 0.96 \qquad (8)$$

Thus, $F1^* = \frac{2*(Precission*Recall)}{Precission+Recall} = 0.8276$.

The obtained metric values demonstrate the potential usability of the proposed method, although there is still room for improvement.

One of the main drawbacks identified in the current implementation is the high rate of false positive $FP$ responses, resulting in a relatively low *Precision* value. This behavior can be attributed to ChatGPT's tendency to attempt to provide an answer even when there is insufficient information available in the given contexts. Additionally, the defined possible intents may not always align perfectly with the provided message, although such intents are often assigned a relatively low probability.

However, it is worth noting that many of these false positive answers were accompanied by true positive $TP$ answers. In other words, while an incorrect answer was provided in some cases, a correct answer was given alongside it. These accompanied false positive answers could be viewed as supplementary information that may be tangentially related to the main answer.

Although the presence of false positives impacts the *Precision* value, the fact that they often coexist with true positives suggests that the system is capable of providing additional insights or related information. This observation highlights the potential value of considering the accompanied false positive responses in a practical context.

Addressing the issue of false positives and refining the alignment between possible intents and the message content are areas for further improvement to enhance the *Precision* of the system.

Let us consider an example. The initial phrase (in Ukrainian) is "На що повинна спиратися ФРМ?" (Eng. – "What should the physical and rehabilitation medicine (PRM) be based on?"). The system detects the following intents:

```
[
    {
        "intent": "subject",
        "type": "interrogation",
        "probability": 0.8,
        "subject": "ФРМ",
        "object": null
    },
    {
        "intent": "cause",
        "type": "narration",
        "probability": 0.6,
        "subject": "ФРМ",
        "object": "спиратися"
    },
    {
        "intent": "way of doing",
        "type": "interrogation",
        "probability": 0.4,
        "subject": null,
        "object": "спиратися"
    }
]
```

And the following named entities were found to select the contexts:

```
[
    {
        "words": ["ФРМ"],
        "type": "noun",
        "main word": "ФРМ"
    },
    {
        "words": ["повинна", "спиратися"],
        "type": "verb",
        "main word": "спиратися"
    }
]
```

We have the following intents that were defined: "subject" (interrogation), "cause" (narration), and "way of doing" (interrogation). All of these intents have relatively high probabilities and should be considered for obtaining the final answer set. However, only the second intent (cause/narration) resulted in the ChatGPT providing a completely correct answer. Surprisingly, this was not the intent with the highest probability. The other intents also led to comprehensive and concise answers, but they were not directly relevant to the initial question. The first intent provided information about physical and rehabilitation medicine (PRM) and the "White Book on Physical and Rehabilitation Medicine in Europe" [44], [45], while the third intent focused on the purposes of the International Classification of Functioning, Disability and Health (ICF) [46], [47]. Although this additional information might be interesting to the user, it is not directly related to the original question.

*Quantitative assessments of the OntoChatGPT information system and ChatGPT*[4]. Here are some key criteria of comparison: response accuracy; knowledge coverage; semantic understanding; conversational quality; knowledge expansion and adaptability (the ability of the systems to expand their knowledge and adapt to new information or domains can be assessed); meta-learning performance.

The quantitative assessment results of OntoChatGPT information system and ChatGPT are given in Table 2.

Please note the limitations:
- the Table 2 provides a general representation of the quantitative assessments based on the specific domain knowledge – rehabilitation medicine – the first chapter of the Ukrainian version of the "White Book on Physical and Rehabilitation Medicine in Europe" [44], [45];
- the free version of ChatGPT can't use extensions via plugins;
- the free version of ChatGPT doesn't have a deep knowledge of the context of external documents (especially for the documents in Ukrainian);

In-depth quantitative comparison of OntoChatGPT and ChatGPT is beyond the scope of this article. The actual quantitative results would depend on the specific evaluation methods, datasets, and performance metrics used in the assessment of OntoChatGPT information system. Throughout the OntoChatGPT system development lifecycle, quantitative results compared to various similar systems will be available in the public GitHub repository associated with this paper.

**Table 2. Quantitative Assessment of OntoChatGPT information system and ChatGPT**

| Metrics | OntoChatGPT | ChatGPT |
|---|---|---|
| Response accuracy | High (82%) | Moderate (70%) |
| Knowledge coverage | Extensive within domains (90%) | General (80%) |
| Semantic understanding | High (90%) | High (90%) |
| Conversational quality | Moderate (70%) | High (above 90%) |
| Knowledge expansion and adaptability | Efficient and adaptable | Limited |
| Meta-Learning Performance | Effective knowledge retention | N/A (not applicable) |

## VI. CONCLUSIONS AND FURTHER PROSPECTIVE

A robust and comprehensive productive triad emerges from this research, encompassing three key components: methodological foundations for utilizing an ontology-driven structured prompts in ChatGPT's meta-learning, advanced information technology, and a composite service known as OntoChatGPT system.

By leveraging ontologies and structured prompts, the OntoChatGPT information system demonstrated its potential for enhancing the performance and knowledge background of ChatGPT in specific subject areas and languages. We formalized the method, which involves meta-learning for creating and tuning structured prompts and the context ontology, and a sequence of operations for detecting intents, identifying named entities, selecting contexts, and forming the final answer prompt based on the information and conclusions to be provided within the contexts. The key feature of this method is its ability to provide ChatGPT with specific information by providing selected contexts, which can enhance its knowledge in specific subject areas and improve its performance with data in different languages.

The use of structured JSON prompts increases their reliability and facilitates obtaining relevant answers. By incorporating the meta-ontology, prompts become more flexible and customizable, taking the meta-learning process to a higher level.

The ontology in the OntoChatGPT information system provides the following impacts:

- *Structured Knowledge.* The ontology provides a structured representation of knowledge, organizing information into concepts, relationships, and properties. It allows for better organization and understanding of the domain-specific information used in the system.
- *Semantic Understanding.* The ontology enables the system to have a deeper understanding of the meaning and context of user queries and prompts. It captures the semantics of the domain, including relationships between concepts, and helps the system interpret and generate more accurate responses.
- *Enhanced Accuracy.* By incorporating domain-specific ontological knowledge, the OntoChatGPT system can improve the accuracy of its responses. The ontology helps the system reason and retrieve relevant information more effectively, leading to more precise and contextually appropriate answers.
- *Knowledge Expansion.* The ontology provides a foundation for knowledge expansion and integration. New information can be added to the ontology, expanding the system's knowledge base and allowing it to handle a wider range of subjects and queries.
- *Domain-Specific Adaptability.* The ontology enables the system to adapt to specific domains or industries by defining domain-specific concepts, properties, and relationships. This allows the OntoChatGPT system to provide more tailored and specialized responses within specific knowledge domains.
- *Interoperability.* The ontology facilitates interoperability by providing a shared understanding of concepts and their relationships. It allows for easier integration with other systems, databases, or ontologies that follow the same or compatible ontological standards.
- *Knowledge-driven Prompts.* The ontology-driven prompts generated by the system use the structured knowledge encoded in the ontology to guide and shape the conversation. These prompts help elicit more specific and relevant information from users and contribute to a more meaningful and productive dialogue.

---

[4] Compared with the free version of ChatGPT available via https://chat.openai.com/

- *Meta-Learning Support.* The ontology provides a meta-learning framework by capturing and organizing knowledge about the learning process itself. It allows the system to learn from user interactions, track performance, and continuously improve its understanding and response generation capabilities.

While the technique has shown promising results, it still exhibits a tendency for false positive answers. However, these false positives are often accompanied by true positive answers and, in cases where relevant information is lacking in the selected contexts, true negative answers. These true negative answers can be seen as providing additional information about the subject matter.

Furthermore, it is important to highlight that the proposed methodology is applicable not only to the specific ChatGPT model used in this study but also to other chatbot systems based on LLM such as Google's Bard, which relies on the PaLM 2 LLM. The underlying principles and techniques of meta-learning, structured prompts, and ontology-driven information retrieval can be adapted and utilized in conjunction with different LLM-based systems. This highlights the potential versatility and scalability of the proposed approach across various chatbot platforms, enabling its wider applicability in the field of natural language processing and dialogue systems.

The proposed approach presented here serves as a preliminary prototype for a more advanced dialogue and reference system that is yet to be developed. In order to enhance the system's performance, several improvements are planned for the structured JSON prompts. Additional inputs, such as sentiments detected in the initial message and the selected contexts, will be incorporated into the prompts. The prompt phrases for keys and values will also be refined, along with the overall structure, to increase their certainty and mitigate the detection of irrelevant intents, thereby reducing the occurrence of false positive answers. These enhancements are expected to improve the Precision criterion value of the system.

## VII. ACKNOWLEDGEMENTS

This study would not have been possible without the financial support of the National Research Foundation of Ukraine. Our work was funded by Grant contract: Development of the cloud-based platform for patient-centered telerehabilitation of oncology patients with mathematical-related modeling [48], application ID: 2021.01/0136.

The research team of the Glushkov Institute of Cybernetics would like to give special recognition to Ellen Cohn (PhD, CCC-SLP, ASHA-F, Department of Communication, University of Pittsburgh, PA, USA), Editor-in-Chief of the International Journal of Telerehabilitation. We greatly appreciate her efforts in promoting Ukrainian science through the dissemination of research works in scholarly publications.

## VIII. DATA AVAILABILITY

The meta-ontology, terms/contexts ontology, SPARQL queries to meta-ontology, samples of structured JSON prompts for the ChatGPT, test questions and results are publicly available via public GitHub repository[5].

Detailed information about the utilization and access to the services included in the OntoChatGPT information system can be obtained upon request. The software alpha version of these services is also available for further exploration and evaluation.

Please reach out to our team to request access and to learn more about the functionalities and features of OntoChatGPT.

---

[5] https://github.com/knowledge-ukraine/OntoChatGPT

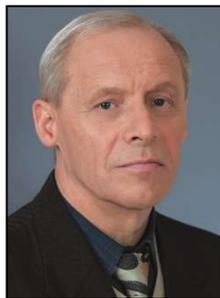

***Oleksandr PALAGIN*** Academician of the National Academy of Sciences of Ukraine, DSc (Doctor of Sciences in Technical Sciences), PhD, Professor, Honored Inventor of Ukraine, Deputy Director for Research of the Glushkov Institute of Cybernetics of the National Academy of Sciences of Ukraine, Head of the Microprocessor Technology Lab. Research interests: AI, Semantic Web; Ontology engineering.

https://orcid.org/0000-0003-3223-1391, palagin_a@ukr.net

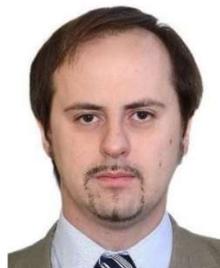

***VLADISLAV KAVERINSKIY*** PhD (technical science), Senior researcher Research interests: AI, Computational linguistics; Phase transformations; Deformational-heat processing; Steels; Aluminum alloys; Powder metallurgy; Metal casting processes.
https://orcid.org/0000-0002-6940-579X, insamhlaithe@gmail.com

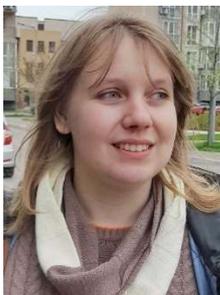

***ANNA LITVIN*** MSc, Junior researcher. Microprocessor Technology Lab, Glushkov Institute of Cybernetics. Research interests: AI, Computational linguistics; Ontology engineering; Dialog systems.
http://orcid.org/0000-0002-5648-9074, litvin_any@ukr.net

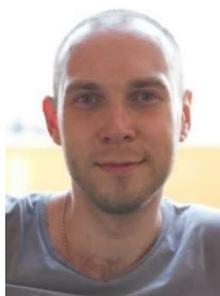

***KYRYLO MALAKHOV*** MSc, Researcher, Backend developer, DevOps engineer. Research interests: AI, Computational linguistics; Ontology engineering; Digital health (Hybrid e-rehabilitation). Member of the expert subgroup on technical issues and architecture of telemedicine within the Interdepartmental Working Group for the development of the concept of implementation of telemedicine in Ukraine.

https://orcid.org/0000-0003-3223-9844, https://linktr.ee/malakhovks, k.malakhov@outlook.com

...